# Training a Probabilistic Graphical Model with Resistive Switching Electronic Synapses

S. Burc Eryilmaz*, *Student Member, IEEE*, Emre Neftci, Siddharth Joshi, *Student Member, IEEE*,
SangBum Kim, *Member, IEEE,* Matthew BrightSky, Hsiang-Lan Lung, Chung Lam, Gert
Cauwenberghs, *Fellow, IEEE,* H.-S. Philip Wong, *Fellow, IEEE*

*Abstract*— Current large scale implementations of deep learning and data mining require thousands of processors, massive amounts of off-chip memory, and consume gigajoules of energy. Emerging memory technologies such as nanoscale two-terminal resistive switching memory devices offer a compact, scalable and low power alternative that permits on-chip co-located processing and memory in fine-grain distributed parallel architecture. Here we report first use of resistive switching memory devices for implementing and training a Restricted Boltzmann Machine (RBM), a generative probabilistic graphical model as a key component for unsupervised learning in deep networks. We experimentally demonstrate a 45-synapse RBM realized with 90 resistive switching phase change memory (PCM) elements trained with a bio-inspired variant of the Contrastive Divergence (CD) algorithm, implementing Hebbian and anti-Hebbian weight updates. The resistive PCM devices show a two-fold to ten-fold reduction in error rate in a missing pixel pattern completion task trained over 30 epochs, compared to untrained case. Measured programming energy consumption is 6.1 nJ per epoch with the resistive switching PCM devices, a factor of ~150 times lower than conventional processor-memory systems. We analyze and discuss the dependence of learning performance on cycle-to-cycle variations as well as number of gradual levels in the PCM analog memory devices.

*Index Terms*—neuromorphic computing, phase change memory, resistive memory, brain-inspired hardware, cognitive computing

## I. INTRODUCTION

Deep learning can extract complex and useful structures within high-dimensional data, without requiring significant amounts of manual feature engineering [1]. It has made significant advances in recent years and is shown to outperform many other machine learning techniques for a variety of tasks such as image recognition, speech recognition, natural language understanding, predicting the effects of mutations in DNA, and reconstructing brain circuits [2]. However, training of large scale deep networks (~$10^9$ synapses, compared to ~$10^{15}$ synapses in human brain) in today's hardware consumes more than 10 gigajoules (estimated) of energy [3-4]. An important origin of this energy consumption is the physical separation of processing and memory, which is exacerbated by the large amounts of data needed for training deep networks [1-5]. It has been reported that ~40 percent of energy consumed in general purpose computers are due to the off-chip memory hierarchy [6], and this fraction will increase when applications are more data-centric [7]. GPUs do not solve this problem, since up to 50 percent of dynamic power and 30 percent of overall power are consumed by off-chip memory as shown in several benchmarks [8]. On-chip SRAM does not solve the problem either, since it is very area inefficient (> 100 $F^2$, F being the minimum half-pitch allowed by the considered lithography) and cannot scale up with system size.

Extracting useful information from data, which requires efficient data mining and (deep) learning algorithms, is becoming increasingly common in consumer products such as smartphones, and is expected to be even more important for the internet-of-things (IoT) [9]; where energy efficiency is especially crucial. To scale up these systems in an energy efficient manner, it is necessary to develop new learning algorithms and hardware architectures that can capitalize on fine-grained on-chip integration of memory with computation.

Because the number of synapses in a neural network far exceeds the number of neurons, we must pay special attention to the power, device density, and wiring of the electronic synapses, for scaled-up systems that solve practical problems. Today, synaptic weights in both conventional processors and neuromorphic processors [10-12] are currently implemented in SRAM and/or DRAM. Due to processing limitations, DRAM needs to be on a separate chip or connected by chip stacking using through-silicon-via (TSV) [13-15] that has limited via density. This results in increased power consumption and limited bandwidth for memory accesses. SRAM, on the other hand, occupies too much area (>100 $F^2$ per bit, 58,800 nm² in 14 nm node CMOS technology [16]) and thus limits the amount of local memory that can be accessed efficiently [11]. "New" non-volatile resistive memory elements [17] such as

This work is supported in part by SONIC, one of six centers of STARnet, a Semiconductor Research Corporation program sponsored by MARCO and DARPA, the NSF Expedition on Computing (Visual Cortex on Silicon, award 1317470), and the member companies of the Stanford Non-Volatile Memory Technology Research Initiative (NMTRI) and the Stanford SystemX Alliance.

S.B. Eryilmaz and H.-S.P. Wong are with the Electrical Engineering Department, Stanford University, Stanford, CA 94305 USA (e-mail: eryilmaz@stanford.edu; hspwong@stanford.edu).

E. Neftci is with the Department of Cognitive Sciences, UC Irvine, Irvine, CA 92697 USA (e-mail: eneftci@uci.edu).

S. Joshi is with the Department of Electrical and Computer Engineering, UC San Diego, San Diego, CA 92093 USA (e-mail: sijoshi@eng.ucsd.edu).

S. Kim, M. BrightSky, C. Lam are with the IBM Research, Yorktown Heights, NY 10598 USA (e-mail: SangBum.Kim@us.ibm.com; breitm@us.ibm.com; clam@us.ibm.com).

H.-L. Lung is with the Macronix International Co., Ltd., Emerging Central Lab, Taiwan (e-mail: Sllung@mxic.com.tw).

G. Cauwenberghs is with the Department of Bioengineering, UC San Diego, San Diego, CA 92093 USA (e-mail: gert@ucsd.edu).

phase change memory (PCM) [18], resistive switching memory (RRAM) [19], conductive bridge memory (CBRAM) [20], and ferroelectric memory (FeRAM) [21] have characteristics that are desirable as electronic synapses. These are two terminal devices with very good size scalability (PCM: 1.2 nm, RRAM: 3 nm, CBRAM: 20 nm demonstrated, corresponding to an area of 5.8 to 400 nm$^2$), low energy programmability (PCM: < 2 pJ, RRAM: 0.4 pJ, CBRAM: 0.7 pJ per bit demonstrated), and analog programmability for implementing synaptic weight within a single device [22, 23]. Monolithic 3-dimensional integration of these nonvolatile memories with CMOS, demonstrated in [24,25], allows designers to hide the logic circuitry underneath multiple layers of synapses, reducing silicon cost and increasing synapse density [4]. Gradual resistance change of these devices has been utilized as a synapse for variety of algorithms [26-33]. These devices can implement variations of biological [18] learning rules within a single device, further suggesting their use for neuromorphic hardware.

Supervised learning refers to finding the right model with labelled data; whereas unsupervised learning refers to fitting the right model to the underlying probability distribution of data while discovering useful features. It is worth noting that human learning is rarely fully supervised, as humans are rarely told the right class of an object or an action, but rather learn by discovering the structure and the context within observations besides the sole category of an object. Unsupervised learning has played a crucial role to revive interest in deep learning [2] and is expected to become far more important in the future due to: 1) exponentially increasing amount of data that is not labelled, 2) inherently unsupervised nature of learning in humans, and 3) improved generalization with unsupervised pre-training observed in experiments [34].

Unsupervised learning using Restricted Boltzmann Machine (RBM) is an important element of deep neural networks for successful generalization especially in environments where huge amount of labelled data is not available [1, 34, 35]. RBMs trained with contrastive learning rules are the backbone of latest research in physical instantiations of deep learning [36]. A probabilistic generative model was recently demonstrated to produce much richer representations of data with less examples than purely supervised deep learning models [37]. RBM is a two-layer probabilistic generative model that represents data in distributive fashion, where one layer consists of visible neurons that is associated with observations, and another layer consists of hidden nodes (neurons) [1]. RBMs provide a very efficient representation of probability distributions in terms of the number of parameters needed [1]. Contrastive Divergence (CD) algorithm [38] is a common technique for training RBMs, and takes advantage of the fact that there is no connectivity between the nodes within layers to perform efficient training. It is important to note that CD learning in RBM is biologically plausible, since neuron models consistent with biological observations can implement CD training in RBMs as well [39].

This paper presents a first demonstration and analysis of CD learning in a RBM with nanoscale electronic synapses. RBM is able to learn 3×3 images and retrieve a missing pixel with more than 80 % probability of success when at most 5 patterns are stored (higher probability of success for less patterns stored). Further improvements in learning performance can be obtained by more precise control of PCM conductance change through device engineering or better programming schemes.

## II. EXPERIMENT AND RESULTS

The fabricated array of PCM devices is shown in Fig. 1(a). The wordline pitch and bitline pitch of the 10×10 array are 1.4 and 1.2 µm, respectively. Phase change elements consist of mushroom GST ($Ge_2Sb_2Te_5$) cells in series with 180 nm CMOS n-channel transistors [40] (Fig. 1(b)). Fig. 1(c) shows that DC switching from high resistance state to low resistance state occurs at around 2 µA. Fig 1(d) shows resistance distribution of cells within a 10×10 array for binary switching. Fig. 1(e),(f) show binary switching and gradual switching behavior, respectively. As we show later, the network is quite robust to device-to-device and cycle-to-cycle variations observed in gradual switching.

Single layer RBM with 9 visible nodes and 5 hidden nodes is implemented on PCM array (Figure 2(a)). Neurons have binary activations in an RBM (0 or 1). RBM parameters encode probability of binary input vectors $v$ as $p(v) = \Sigma_h p(v,h) = (\Sigma_h e^{-E(v,h)})/Z$ where $E(v,h) = -\Sigma_{i \in visible} a_i \times v_i - \Sigma_{j \in hidden} b_j \times h_j - \Sigma_{i,j} w_{ij} \times v_i \times h_j$ is the 'energy' of the network for a $(v,h)$ visible and hidden vector pair, and $p(v,h) = e^{-E(v,h)} \times 1/Z$ is the probability assigned to a $(v,h)$ pair by the model. $Z$ is the normalizing parameter for realizing $\Sigma_h [\Sigma_v p(v,h)]=1$.

Nine input neurons represent each pixel in a 3×3 image for the task we performed in this work. Since SET and RESET rates are very asymmetrical for PCM devices, learning cannot be efficiently realized with single cell synapses [29]. Hence, we chose to use 2 PCM devices per synaptic weight [29] (see

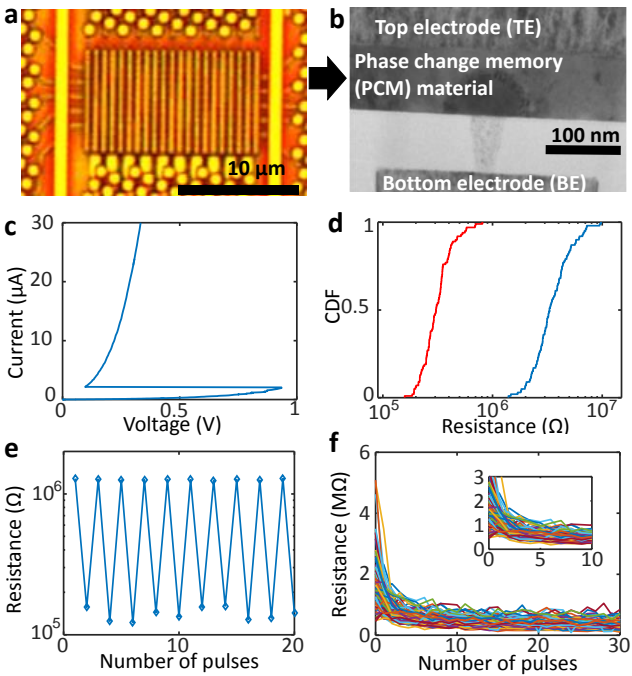

Fig. 1. (a) Chip micrograph of crossbar PCM array (b) TEM cross section of a PCM device. (c) DC SET switching of a single device (d) Cumulative distribution (CDF) of high and low resistance values in an array (e) Binary switching of a PCM device (f) Gradual SET behavior for 100 different devices. Inset shows the zoomed-in version of the same data.

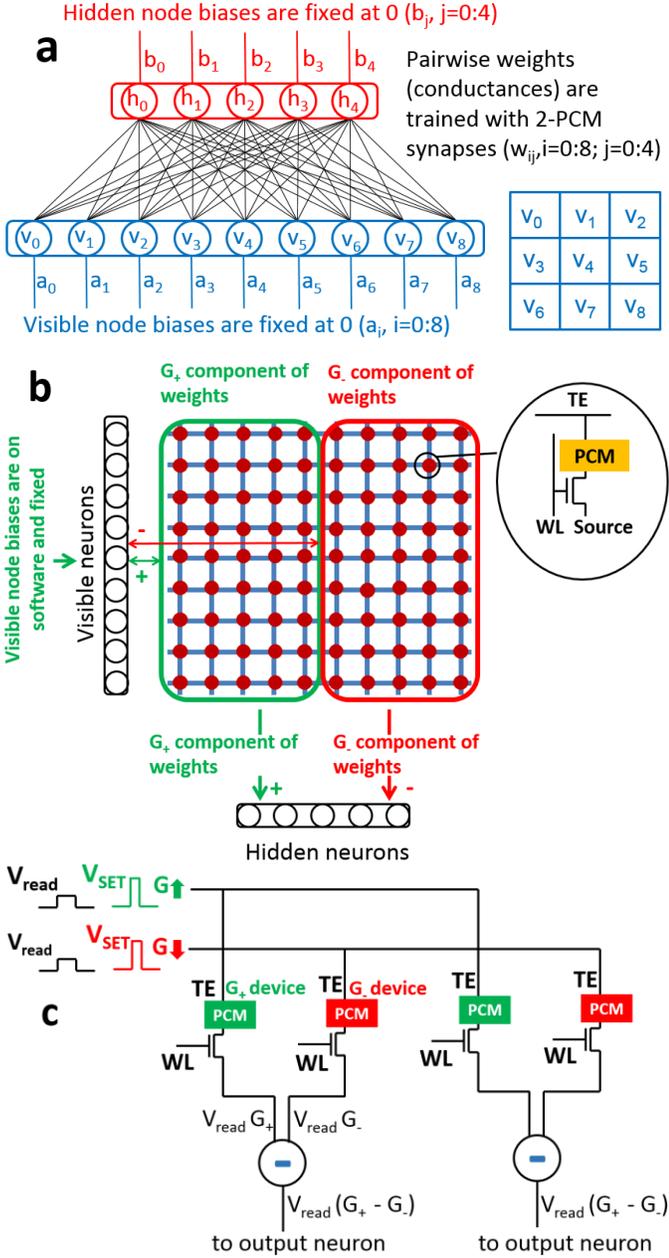

Fig. 2. (a) Architecture of the RBM with pairwise weights between visible and hidden units and single node biases of visible and hidden units. 9 visible neurons represent binary image pixel values. 5 hidden neurons are used to learn the features. Visible and hidden node biases are fixed, and pairwise weights are learned. (b) Mapping of weights ($w_{ij}$) on PCM hardware. (c) Diagram of 2-PCM synapse implemented with 2 memory cells.

Fig. 2(b)). Parameters consist of single-node biases for visible (9 parameters) and hidden (5 parameters) nodes, as well as pairwise synaptic weights (45 parameters), as shown in Fig. 2(a), adding up to 59 parameters used to represent the probability distribution of input data that consists of binary vectors with length 9.

In order to fit the parameters on the 10×10 array, we only treat pairwise weights as variables. We fix visible and hidden node biases at 0 and do not train them. 0 bias corresponds to the assumption that each node is equally likely to be ON or OFF with no information from other nodes, which is indeed true for visible nodes with the dataset in Fig. 3(b) (overall, each pixel is ON half of the time). This results in 45 parameters treated as variables, represented by 90 PCM devices from the 10×10 array. Those parameters are mapped on a PCM array as shown in Fig. 2(b). According to Fig. 2(c), the effective conductance of each weight is given by G=($G_+$ - $G_-$), where $G_+$ and $G_-$ are the conductances of each of the two differential PCM cells within the 2-PCM synapse. Weights are mapped on the network as $w_{ij}$=($G_{ij}$-M)/S where G is the effective synaptic conductance as described above, and the constants M and S are the mean and standard deviation of initial resistance values. M and S are determined after initialization phase, and stays the same during training and for the whole network (see Table I). The initialization procedure starts with applying a strong RESET pulse for each PCM device. After initialization, training proceeds as described in Fig. 3, using 3×3 images in the bars-and-stripes dataset in Fig. 3(b) as input [41].

The CD algorithm updates the parameters $w_{ij}$ as follows to maximize data log likelihood: $\Delta w_{ij} \propto <v_i h_j>_{data} - <v_i h_j>_{model}$ (see Fig. 3(a)) [1,38]. Here, $<v_i h_j>_{data}$ is the average (over the data) of pairwise interaction of activations of each visible and hidden neuron, and $<v_i h_j>_{model}$ is the average of the same quantity, but taken over the probability distribution represented by the model during that iteration before the weight update. A cycle that consists of updating the weights after presenting all images (or a subset of images selected as dataset) in Fig. 3(b) is called an epoch, or iteration. In one iteration, the term $<v_i h_j>_{data}$ is calculated as shown in Fig. 3(a) and Table 1: $v_i h_j$ is calculated for each data vector by sampling hidden neurons and is averaged for all dataset. This is done by measuring the total current through the source node for each hidden neuron when the corresponding read voltage (0.1 V) is applied at the bit-line (depending on which visible nodes are ON) and corresponding word-line (depending on which hidden neuron input is being measured). Output of each hidden neuron is then 1 with a probability given in Fig. 3(a), and 0 otherwise. Exact calculation of the term $<v_i h_j>_{model}$ is

TABLE I
PSEUDOCODE FOR THE ALGORITHM USED IN LEARNING AND INFERENCE

```
RESET all devices; //initialization
visible_biases, hidden_biases = 0;
G+ = conductance(G+ devices);
G- = conductance(G- devices);
M = mean(G+ - G-); S = stddev(G+ - G-);
for iteration = 1:maxIter
  for i = 1:Nv  //Nv = 9 is visible layer size
    for j = 1:Nh //Nh = 5 is hidden layer size
      wij = [(G+ij – G-ij) – M] / S;  //weights
      (vihj)data = 0; (vihj)model = 0; //initialization
  for index = 1:dataSize
    v = dataset[index];
    for i = 1:Nv
      for j = 1:Nh
        [(vihj)data]index = vihj;
    [(vihj)model]index = run_gibbs_sampling(v);
  for i = 1:Nv
    for j = 1:Nh
      <vihj>data = average_over_data[(vihj)data];
      <vihj>model = average_over_data[(vihj)model];
      Δwij = <vihj>data – <vihj>model;
      if Δwij > 0
        partial_SET(device G+ij);
      else
        partial_SET(device G-ij);
```

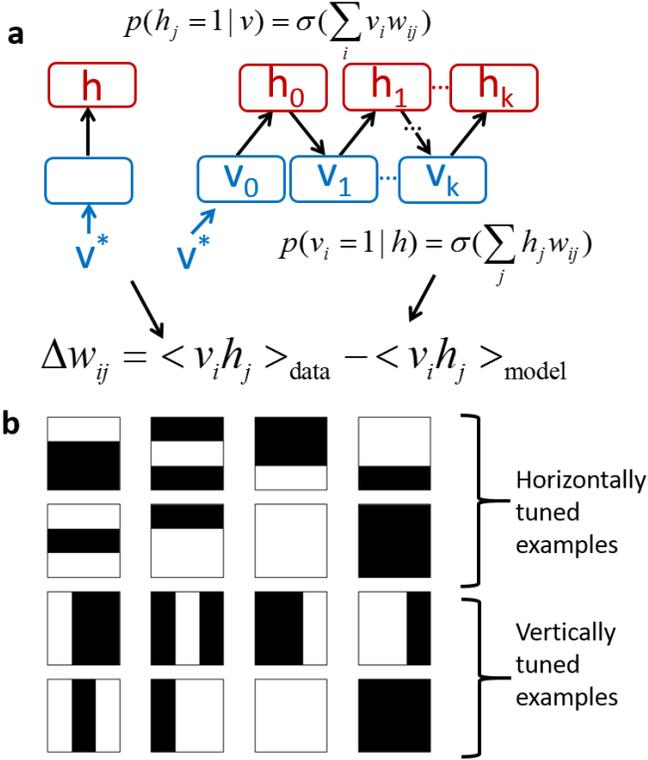

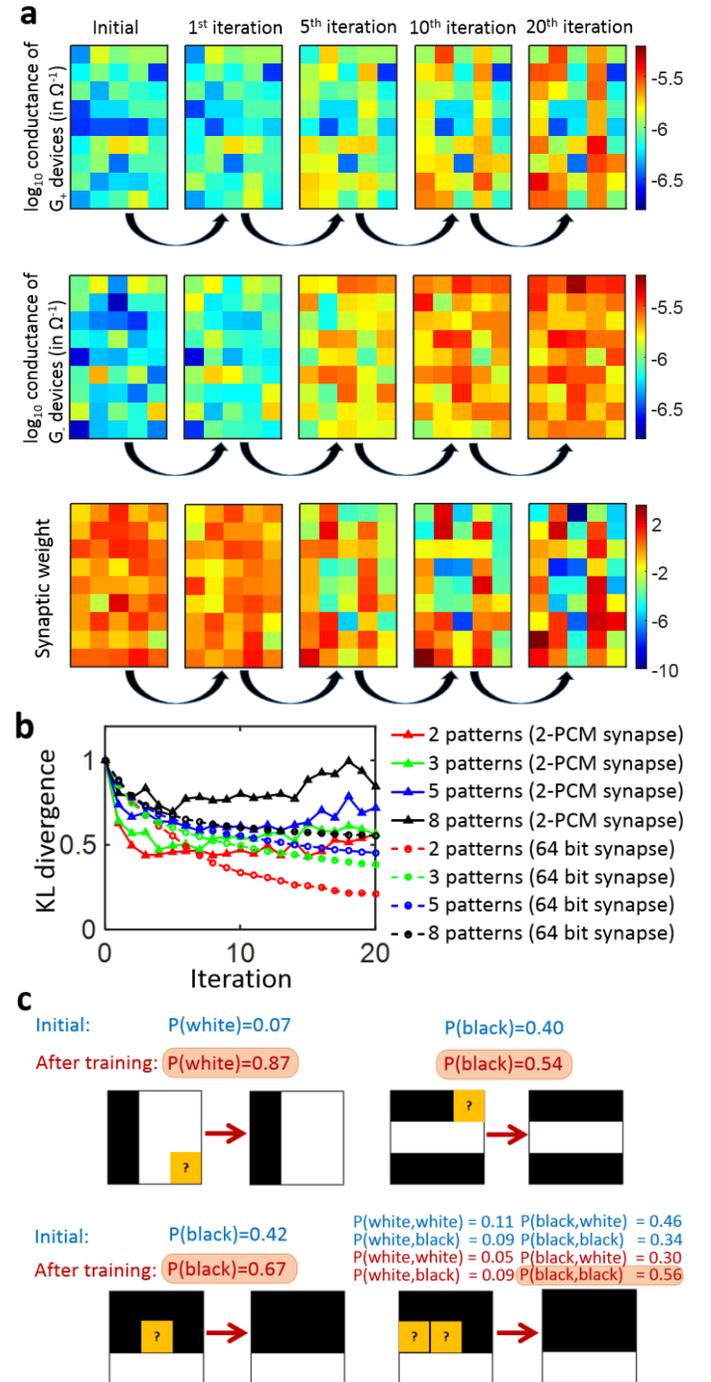

Fig. 3. (a) Computing the first and second terms of weight update equation. σ is the sigmoid function, and v* represents the input data used to initialize visible units. The first term can be easily computed from data. The second term requires Gibbs sampling. (b) 'Bars and stripes' dataset [41] used in this experiment. Data are generated by randomly selecting either horizontal or vertical tuning with equal probability for a 3×3 image, and then correspondingly turning each row or column ON (white) or OFF (black) with equal probability.

intractable, so approximation is made by Gibbs sampling, which is shown to be very efficient for RBM and empirically give good results with CD training [1,38]. For Gibbs sampling, input image is presented, and top-down iterations are performed k-times (we use k=3) where the input is removed after it is presented at the beginning [1,38]. During these iterations, visible node outputs (from top to bottom) and hidden node values (from bottom to top) are calculated probabilistically, using the conductance values of PCM devices, as shown in Fig. 3(a). Calculation of $<v_i h_j>_{model}$ and $<v_i h_j>_{data}$ is performed by software on a computer, while the PCM hardware serves as hardware synapses. Once the terms $<v_i h_j>_{data}$ and $<v_i h_j>_{model}$ are calculated, depending on which term is bigger for each synaptic weight, either $G_+$ or $G_-$ device is applied gradual SET pulse (see Fig. 2(c) and Table I). This means there is a total of 45 gradual SET operations in one weight update phase.

We monitor the training procedure by tracking the Kullback-Leibler (KL) divergence [42] between the probability distribution of the dataset and the probability distribution represented by the network at every iteration. KL divergence is a similarity metric between two probability distributions, and smaller KL divergence means that the probability distribution represented by the network can model the probability distribution of the original data more accurately. Computing KL divergence exactly is in general intractable, but can be efficiently and relatively accurately approximated using annealed importance sampling (AIS) [43], and we use AIS here to estimate the KL divergence. Fig. 4(a) shows that the evolution of PCM conductances starts to saturate after the 10th epoch, and only minor changes to conductances are observed from 10th until the 20th epoch. This causes the KL divergence to improve overall until the 10th epoch. Towards the 10th epoch, the improvement slows down, and towards the 20th epoch the progress starts reverting, more severely when the dataset (the patterns stored in network with training) contains larger number of distinct patterns. The fluctuations in KL divergence starting towards 10th epoch and the accompanying slowdown of improvement in KL divergence can be explained by the observation that for a given device, the gradual resistance change starts to become small compared to the cycle-to-cycle variation in Fig. 1(f). The onset of unlearning (increase in KL divergence) towards 20th epoch happens when for sufficiently large number of synapses, either the $G_+$ or the $G_-$ device saturates. When $G_+$ saturates, for instance, in the subsequent iterations, it cannot respond to the change in $G_-$, degrading the learning feedback loop and causing 'unlearning' ('forgetting') as training progresses further.

Fig. 4. Results for the case where 5 patterns are stored. (a) Evolution of PCM conductances and weights during training. Weights are mapped as [($G_+$ - $G_-$)-M]/S (see text and Table 1 for definition of S and M). (b) KL divergence vs number of training epochs for PCM synaptic array measured in hardware for different cases, and 64-bit digital synapse array emulated in software on a conventional computer. Data points are averaged over 5 trials for each case. (c) Inference results for sample images with missing pixels, showing pattern completion after training. For instance, P(white) = 0.87 means the network assigns a probability of 0.87 to white and 0.13 to black for a given pixel.

TABLE II. EXPERIMENTAL ARRAY LEVEL LEARNING DEMONSTRATIONS COMPARED WITH THIS WORK

| | This work | Prezioso et al. [31] | Park et al. [27] | Burr et al. [30] | Kim et al. [33] | Eryilmaz et al. [26] |
|---|---|---|---|---|---|---|
| Network type | Probabilistic graphical model (RBM) | Single-layer perceptron | Single-layer perceptron | Multi-layer perceptron | Auto-associator | Auto-associator |
| Learning algorithm | Contrastive Divergence (Hebbian + anti-Hebbian) | Backpropagation (Delta rule) | Hebbian | Backpropagation | Hebbian | Hebbian |
| Number of synapses | 45 | 30 | 192 | 164,885 | 65,536 | 100 |
| Dataset | Bars&stripes | Binary patterns | Processed EEG data | MNIST dataset | Binary patterns | Binary patters |
| Input size | 9 pixels (3x3) | 9 pixels (3x3) | 32-bit binary vector | 528 pixels (22x24) | 256 pixels (16x16) | 10 pixels (5x2) |
| Number of patterns learnt | 2-16 patterns | 3 patterns | 3 classes | 10 classes | 9 patterns | 2 patterns |
| Supervision | Unsupervised | Supervised | Supervised | Supervised | Unsupervised | Unsupervised |
| Success rate | ~ 98-55% | 100% after 20-30 epochs | NR | 82-83% | NR | 100% after 1-11 epochs |
| Energy or power reported | ~ 6.1 nJ/epoch (synapses only) | NR | 47.9 mW (neurons and synapses) | NR | NR | ~ 4.8 nJ/epoch |
| Neurons | Off-chip | Off-chip | Off-chip | Off-chip | On-chip | Off-chip |
| Resistive memory used | Phase change | Metal-oxide | Metal-oxide | Phase change | Phase change | Phase change |

The 2-PCM synapse hardware implementation is compared to a computer simulation of the CD algorithm where the synaptic weight is stored in the form of double-precision floating point numbers (labeled as "64-bit synapse"). It is interesting to note that 2-PCM synapse results in better learning until 10th epoch, but the 64-bit synapse results in lower KL divergence after the 10th epoch, as shown in Fig. 4(b). This is due to the onset of saturation effect within PCM devices described above, whereas 64-bit synapse has infinite (for practical purposes) precision and does not saturate. Fig. 4(c) shows some inference tests where the probability of the value of missing pixel estimated by the network before and after training is given for some cases. In all these cases, the RBM performs significantly better after training compared to before training, predicting the right combinations after training. Note that even when 2 pixels are missing (see Fig. 4(c)), the network can predict the right combination among 3 other combinations after training, while its prediction was wrong before training. We measure ~3.2 nJ/epoch total energy consumed during training within the PCM devices in the array. This is much less than the energy consumption estimated for a conventional computer performing the same task, as described later.

We further study the relationship between the number of iterations and the number of patterns stored in the network on the correct recovery of a missing pixel (error rate). Analysis of error rate is more applicable to practical cases than KL divergence, although KL divergence gives a good indication of training performance. We performed 5 trials for each of different cases where different number of patterns are stored in the network. During inference, all cases of 1 missing pixel are evaluated (9 cases for 1 trial) and the probabilities are averaged. For each case, the obtained probabilities are averaged. For each number of stored patterns and for each trial, a subset of training set shown in Fig. 3(b) is randomly selected with the number of elements of the subset being equal to the number of stored patterns. As Fig. 5 shows, the 2-PCM synapse gives better performance for most cases until the 30th epoch. Performance for 2-PCM synapse case slows down towards 30th epoch, and after 70th epoch, recovery error rate becomes larger than that after 10th epoch. Better performance of PCM synapse until 30th epoch is due to the nonlinearity in the change of resistance of PCM. For both cases, we used the same learning rate for weight update equations; and it is worth noting that it is possible to find an optimum learning rate for 64-bit synapse such that it starts performing better than 2-PCM synapse before 30th epoch. Recovery error rate for 64-bit synapse, although worse than 2-PCM synapse for the first 30 epochs in our experiment, continues improving until 70th epoch, and becomes better than the best case of 2-PCM synapse. The 2-PCM synapse is preferable for applications where some performance can be traded off in favor of huge energy savings (see Discussion section).

III. DISCUSSION

The fluctuations in Fig. 4(b) for the 2-PCM synapse case compared to 64-bit synapse case can be attributed to (a) limited control over the conductance change of PCM devices from cycle-to-cycle, and (b) implementing signed constant-amplitude updates rather than the full gradient in Fig. 3(a). It is expected that achieving more gradual levels through device engineering [44-46] or a different programming scheme [44] can lead to better results in training. A better control of gradual conductance change also helps mitigate the impact of device-to-device variations during training, since the weight updates would take care of this type of variation through changing the weight accordingly [4]. In fact, device-to-device variations intrinsically allow symmetry breaking at the initialization stage which is needed for initializing weights in neural networks before training [1]. For conventional hardware, this is done by initializing the weights by assigning a value from Gaussian distribution with mean 0. This is much better for learning than initializing all weights to 0. On the other hand, very large device-to-device variations at the initialization stage can be harmful for convergence. In this experiment, we scaled the conductances by the standard variation to obtain the weight values (see Table I) to avoid very large resistance deviations at the initialization. This can be implemented on circuit level either by using an appropriate voltage level for read pulses or configurable neuron parameters.

6Energy consumed by the wires is only a few pJ, so it is negligible compared to 3.2 nJ/epoch programming energy consumed within PCM. However, wire energy is important for larger arrays, hence should be considered when scaling up in system size [22]. For a 1k × 1k array, wires consume from 30 pJ for 20 nm array half pitch [22]. For comparison, our devices consume ~72 pJ in a partial SET for this experiment. To compare the energy efficiency of our approach with a conventional hardware, we estimate synaptic energy consumption for the same training procedure that is run on conventional hardware. For a direct comparison, we only compare the energy consumption within the synaptic weights, and we do not include the energy consumption due to computation within the neurons. We assume 5 patterns are stored during training, and we measure the average energy consumption over 5 trials with PCM hardware. Using PCM hardware, average energy consumption due to programming is 3.2 nJ/epoch. Energy consumption due to read is much larger since the integration time of our equipment is 80 ms. For a fair comparison, we assume 50 µs integration time, which is the specification of a commercially available current-input ADC with 20 bit resolution [47]. Note that we do not include power consumption of the ADC since the optimum design of neuron circuits and periphery is outside the scope of this paper. This results in 2.9 nJ/epoch energy consumption due to read on average (see Appendix for details). Read energy also depends on the resistance distribution of PCM cells across the array, higher resistance being more favorable. This can be observed in Supplementary Figure 1, where read energy/epoch increases over time as the device resistances decrease due to gradual SET operations. Overall, training and inference results in 6.1 nJ/epoch energy consumed within PCM cells. This task corresponds to the following for a conventional hardware: 1) Read weights from non-volatile storage (here assumed on-chip as described below) into processor (assume no DRAM) 2) Perform 7 matrix-vector multiplications to perform KL divergence shown in Fig. 3(a) 3) Perform matrix addition for weight updates 3) Write back the result into non-volatile storage. Note that in a realistic scenario, we might not update weights very often, hence keeping the updated weights in non-volatile storage is desired. For a fair comparison, we assume PCM that is monolithically integrated on top of processor for non-volatile memory access, which is recently proposed as the next-generation hardware for data-abundant applications [6]. With the characteristics of phase change memory used in our experiment, and reported energy consumption of vector operations in Intel Xeon Phi processor (in 22 nm technology node), we estimate 910 nJ energy for the equivalent synaptic operations performed on PCM in our experiment (430 nJ within logic, 480 nJ for memory access; not accounting for energy consumption of the memory access circuits and sense amplifiers for a fair comparison). Hence, our hardware consumes ~150× lower energy per epoch on average, compared to next-generation conventional hardware with on-chip digital memory. Assuming 16 bit synapses for digital conventional hardware and assuming 16-bit ADC used in neuromorphic PCM hardware (20 µs read time [48]) gives 230 nJ for conventional hardware and 4.4 nJ for PCM hardware for synaptic operations. We further estimate energy consumption for PCM-based neuromorphic hardware and conventional hardware using data obtained from large scale PCM array measurements in the literature. An estimate using characteristics of a 1 Gb PCM array in 45 nm node [49] results in 590 nJ for conventional hardware, and 19 nJ for PCM hardware, translating to >30× energy savings in favor of PCM neuromorphic hardware. Big energy savings here come from tightly coupled memory and processing, and avoiding data movements between memory and processing elements. Details of these estimates are given in Appendix. These estimates should provide a perspective for comparison between two-terminal resistive switching non-volatile memory based neuromorphic hardware and conventional hardware. Table 2 shows comparison of this work with previous demonstrations of learning using resistive memory based synapses. Compared to previous works, our approach is unique in that it combines: 1) Representational power of probabilistic graphical models and 2) Biologically plausible learning rule. Although the prototype shown in this work is small with respect to targeted applications, it can be scaled to larger sizes if the saturation effect can be mitigated. Apart from solutions that involves device engineering or modification of programming scheme as discussed above, refresh procedure can be used to mitigate the saturation effect in phase change memory [30]. This method requires that resistance values should be refreshed periodically to high-resistance regime, while keeping the synaptic strength the same within some margin. Studies have shown that this method mitigates saturation effect with sufficiently low refresh periods [30], although more experimental data and studies on the overhead of refresh operation are needed. This refresh scheme might not be necessary if the device can exhibit bidirectional gradual resistance change as opposed to unidirectional resistance change observed in this experiment.

In this work, the quantity $R_{off}/R_{on}$ is between 10-100 as observed in Fig. 1(f) under the pulsing conditions used in the experiment. It was reported elsewhere that when 2 cells are used in differential mode to implement a synaptic weight as in this work, $R_{off}/R_{on}$ requirement relaxes substantially since the dynamic range of synaptic weights ($G_+ - G_-$) can cover very small weights (as small as the precision of gradual resistance change allows) without the need to have $R_{off}$ to be much bigger than $R_{on}$ [4]. It is important to note that when synaptic weights consist of 2 memory cells in differential mode, $R_{off}/R_{on}$ needed is a function of 1) the precision of the control of gradual resistance change and 2) the absolute value of the ratio of largest and smallest weights for the network that can model the data distribution well. For larger networks, the second item might have a different requirement, but a ratio of 500 for $R_{off}/R_{on}$ was reported to give good results in large scale network simulations with realistic memory device models [4]. It is also possible that the finite $R_{off}/R_{on}$ might act as a regularizer during training by limiting the dynamic range and can be useful for better generalization after learning [1], but this requires further analysis of device-algorithm interactions.

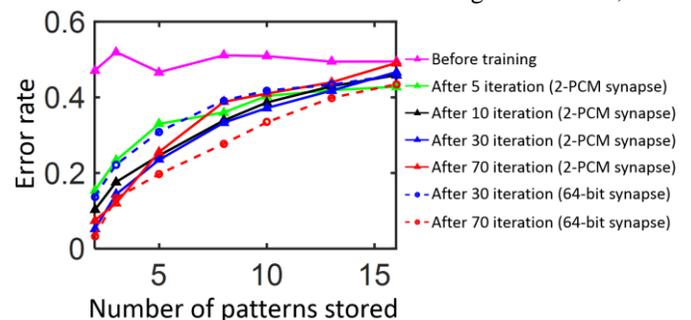

Fig. 5. Probability of incorrect recovery (error rate) of a missing pixel vs. number of stored patterns. Results are averaged over 5 trials for each case.

In the small scale array used in our experiment, yield was not an issue and we observed full yield, mainly because 1) the small size of the array 2) devices that would be considered as failed to be used for binary memory due to very low $R_{off}/R_{on}$ ratio can still function as an analog memory device for the purposes of learning, to the extent that it can exhibit sufficiently good gradual resistance change behavior. We expect yield to be a more important issue for practical larger scale networks, and more studies regarding the effect of yield on learning performance are needed.

While in this work we evaluate the performance of RBM individually, RBM is very commonly used as a pre-training method for deep networks, where weights are fine-tuned using backpropagation algorithm following RBM training [35]. For this type of application, the performance of RBM should be evaluated together with the backpropagation phase following RBM training. It is possible that performance requirements for RBM training might be relaxed for this type of application. Depending on how the PCM devices perform with backpropagation algorithm [35], device requirements might change depending on overall training performance with RBM pre-training and backpropagation-based fine tuning combined.

IV. CONCLUSION

We experimentally demonstrate a proof-of-concept implementation of a probabilistic graphical model (RBM) using 45 synapses implemented with 90 phase change memory (PCM) elements trained with CD algorithm. We observe and monitor learning through KL divergence as well as inference tests when 1 to 2 pixels are missing and error rate of recovery when a pixel is missing. Synaptic operations consume 6.1 nJ/epoch, compared to an estimated 910 nJ/epoch for a state-of-the-art conventional processor. Fluctuations in the learning progress is observed when conductance change control is not good enough to overcome cycle-to-cycle variations. This work reveals the opportunities in utilizing emerging non-volatile-memory devices for probabilistic computing, and provides useful guidelines for refining the programming schemes and device characteristics for efficient learning.

APPENDIX

All measurement equipment is controlled by a Python interface on a PC, which is also used to perform software operations (see Supplementary Figure 2). In all the measurements, the resistance of the memory cell is measured by applying 0.1 V read voltage at the bit-line and 3.3 V at the word-line. For binary switching measurement in Fig. 1(d) and (e), alternating SET pulses (1.5 V amplitude, 100 ns/800 ns/2µs rise/width/fall time) and RESET pulses (2.5 V amplitude, 5 ns/50 ns/5 ns rise/width/fall time) are applied by pulse generator at the word-line, and a wider pulse that overlaps SET and RESET pulses in time is applied on the bit-line (10/1000/10 µs rise/width/fall time, 3.3 V amplitude). In all measurements involving gradual SET (Fig 1(f) and learning phase during the experiment), a gradual SET pulse (10 ns/50 ns/10 ns rise/width/fall time, 1.1 V amplitude) is applied at the word-line, and a wider bit-line pulse is applied as below. RESET pulse with the same specifications above is applied to initialize the resistance values of all PCM cells in the 10×10 array.

We estimate energy consumption for a similar task implemented on a Xeon Phi processor using the characterization and modeling results given in ref. [50]. In contrastive divergence, a pass from visible to hidden nodes is a multiplication of two matrices A×B where A has rows equal to number of data vectors stored (5 in the example), and with columns equal to number of visible neurons (9); and B has number of rows equal to number of visible neurons (9) and number of columns equal to number of hidden neurons (5). This corresponds to 73.125 vector operations on average, which results in 73.125 nJ energy consumption where each vector operation consumes 1 nJ [50]. A pass from hidden layer to visible layer is the multiplication of two matrices C×D where C has number of rows equal to number of data vectors in dataset (5), and number of columns equal to number of hidden neurons (5); and D has number of rows equal to number of hidden neurons (5), and number of columns equal to number of visible neurons (9). This results in 45 vector operations on average, which consumes 45 nJ. For simplicity, we assume a vector addition consumes the same amount of energy with vector multiplication in Intel Xeon Phi, since [50] only reports the energy for vector addition and not multiplication. We further assume that synaptic weights in conventional hardware are 64-bit double precision floating points. Overall, 1 iteration of CD (contrastive divergence) update has 4 visible–to-hidden pass and 3 hidden-to-visible passes, resulting in 427.5 nJ energy consumption. At the end, weight update is performed by adding ∆W to W (W is the weight matrix), which is another 6 vector addition, consuming 6 nJ. For estimates using 1 Gb PCM array [49], we extract the SET and RESET voltage and current from the figures in [49] as 1.8 V/100 µA for SET and 2.2 V/200 µA for RESET for calculating energy due to SET and RESET. Since the pulse width for programming pulses is not reported in [49], we assume a RESET time of 50 ns, SET time of 400 ns (taken from [51]), and read time of 20 ns for calculating energy values. For calculating mean conductance to use in read energy calculation, we extract $R_{low}$ and $R_{high}$ to be 10 kΩ and 2 MΩ, respectively, and used the equation $(1/R_{low} + 1/R_{high})/2$ for calculating mean conductance.